*Type of the Paper-Article*

# SMOTE-ENC: A novel SMOTE-based method to generate synthetic data for nominal and continuous features

Mimi Mukherjee [1], Matloob Khushi [2,*]

1 School of Computer science, University of Sydney, Sydney, NSW, Australia; mimi_mukherjee@outlook.com
2 School of Computer science, University of Sydney, Sydney, NSW, Australia; mkhushi@uni.sydney.edu.au
* Correspondence: mkhushi@uni.sydney.edu.au

**Abstract:** Real-world datasets are heavily skewed where some classes are significantly outnumbered by the other classes. In these situations, machine learning algorithms fail to achieve substantial efficacy while predicting these under-represented instances. To solve this problem, many variations of synthetic minority over-sampling methods (SMOTE) have been proposed to balance the dataset which deals with continuous features. However, for datasets with both nominal and continuous features, SMOTE-NC is the only SMOTE-based over-sampling technique to balance the data. In this paper, we present a novel minority over-sampling method, SMOTE-ENC (SMOTE – Encoded Nominal and Continuous), in which, nominal features are encoded as numeric values and difference between two such numeric value reflects the amount of change of association with minority class. Our experiments show that, classification model using SMOTE-ENC method offers better prediction than model using SMOTE-NC when the dataset has a substantial number of nominal features and also when there is some association between the categorical features and the target class. Additionally, our proposed method addressed one of the major limitations of SMOTE-NC algorithm. SMOTE-NC can be applied only on mixed datasets that have features consisting of both continuous and nominal features and cannot function if all the features of the dataset are nominal. Our novel method has been generalized to be applied on both mixed datasets and on nominal only datasets. Code is available from https://mkhushi.github.io/.

## 1. Introduction

Class imbalance is one of the major obstacles in classification problems to achieve high accuracy for minority class data points. In real-world, datasets are very often heavily skewed; e.g., in medical datasets [1] for cancer detection, benign tumors significantly outnumber cancerous tumors [2-7], in financial datasets [8], legitimate transactions outnumber fraudulent transactions [9]. When statistical models are constructed on these extremely skewed datasets, model tends to predict every instance as a member of over-represented class, making the model inefficient to identify under-represented class instances, which is especially critical when classifying minority instances is the point of interest.

In order to solve this problem, in machine learning algorithms, different approaches are adopted, which, broadly, can be divided into three categories: tuning cost function[10], sampling[11] and modifying learning algorithm[12]. Among these, sampling techniques are one of the widely accepted method to balance the skewed dataset. One of the earliest proposed sampling method was random under-sampling[13] and random-over-sampling: the former samples majority class data points to form a representative set similar size as the under-represented class and discards the rest while the latter replicates some of the minority class data points to balance the dataset. However, while by adding additional copies to some minority class instance, random over-sampling leads to overfitting and by removing samples removes potentially valuable data [14], random under-sampling causes underfitting. Hence, both of them exhibit poor generalization on unseen data.

As an alternative way, in 2002, Chawla et al. [15] proposed a novel over-sampling method SMOTE, where, instead of replicating existing instances, new instances were synthesized to balance the dataset. In SMOTE, for each under-represented instance, a predetermined number of neighbors were calculated, then some minority class instances were randomly chosen for synthetic data point creation. Finally, artificial observations were created along the line between the selected minority instance and its closest neighbors. Chawla et al. experimented SMOTE on a wide range of datasets which showed that SMOTE can improve classifier performance for minority class. By virtue of being trained on more under-represented examples SMOTE decreased generalization error.

Since then, different variations of SMOTE have been proposed [16-21]. Based on SMOTE method, Han et al. proposed BorderLine-SMOTE, [22] where minority instances close to the class boundary were synthesized. In 2004, Batista et al.[11] proposed SMOTE-ENN and SMOTE-Tomek which considered distribution overlap and class boundaries while generating synthetic data. In 2008, He et al. proposed ADASYN[23] which focused on difficult minority class instances and simulated synthetic data by understanding pattern of those instances. In 2016, Torres et al. proposed a variant of SMOTE, SMOTE-D[24], which, unlike SMOTE, deterministically synthesized artificial data points for under-represented class. All of these methods were developed considering continuous variables and when applied on datasets with categorical features, failed to identify nominal features which resulted in creating new labels for these attributes.

To solve this problem, Chawla et al. proposed a method SMOTE-NC [15] (SMOTE Nominal-Continuous) which treated nominal attributes different than continuous attributes and preserved the original labels of categorical features in the resampled data. However, for multi-label nominal features SMOTE-NC failed to interpret the difference in association between each label and minority class target. Also, SMOTE-NC method can only function when the dataset has at least one continuous attribute.

In this paper, we propose a novel algorithm, SMOTE-ENC, which, irrespective of presence of continuous feature, will comprehend that different labels of nominal features have different affinity towards minority class. We compare this new algorithm with SMOTE-NC and our experiments indicates that this new method performs better than SMOTE-NC for datasets with numerous multi-label nominal features.

## 2. Materials and Methods

Since, SMOTE method is based on k-nearest neighbor algorithm, distance calculation between two instances is the most important aspect of generating synthetic samples. In the feature space, deriving distance between continuous variables are straight-forward, however for categorical variables calculating distance between two labels can be complex. In SMOTE-NC, if label of a categorical attribute differs between an instance and its nearest neighbors, then a fixed-value, i.e., median of standard deviation of continuous features, is added in the distance calculation. Hence, in this method, for a multi-label categorical feature, distance between any two labels is same.

Now, let's consider a dataset where target is to identify a potential buyer of life-insurance plan. Suppose, the dataset has a nominal feature, marital status, with three labels – single, married and divorced and following is the percentage distribution of people who bought life insurance plan. Single –> 18%, divorced –> 22% and married –> 60%. In SMOTE-NC, distance between married - divorced and divorced - single is considered to be same. However, based on the data, it seems people who are divorced or single, are less interested to buy life insurance plan than married people. So, considering the similar affinity toward target, single and divorced labels should be closer to each other than to married label, i.e., distance between married-divorced should be more than single-divorced. In SMOTE-ENC, labels of a categorical variable are represented by their affinity of association with minority class target.

SMOTE-ENC algorithm

Input: t = number of minority class samples in training set; n% = amount of oversampling; k = number of nearest neighbors to be considered, s = Total number of instances in training set, c = Number of continous variables in the dataset, m = median of standard deviation of continuous features when c > 0

ir = t/s
for each categorical feature, do
    for each 'l' in distinct labels, do
        e = total number of 'l' labelled instances in training set
        e` = $e * ir$;
        o = number of 'l' labelled minority class instances in training set
        $\chi = \frac{(o-e`)}{e`}$;

        if c > 0,
            l = $\chi * m$;
        else,
            l = $\chi$;
    end
end
Apply SMOTE (t, n, k)
For the synthetic data points, categorical attribute's value is decided as the value in the majority of its nearest neighbors
Inverse-encode categorical values to their original labels

(1)

We explain our algorithm with a sample training dataset (Table 1) with three continuous features (C1, C2, C3) and a categorical feature (N). The dataset is imbalance where we have 2 instances minority class *min* and 3 instances of majority class *maj*.

**Table 1.** Sample dataset to explain our algorithm (SMOTE-ENC)

|  | C1 | C2 | C3 | N | target |
|---|---|---|---|---|---|
| Instance 1 (i1) | 100 | 20 | 85 | a | *min* |
| Instance 2 (i2) | 200 | 65 | 54 | b | *min* |
| Instance 3 (i3) | 166 | 24 | 38 | a | *maj* |
| Instance 4 (i4) | 344 | 67 | 89 | b | *maj* |
| Instance 5 (i5) | 200 | 30 | 75 | b | *maj* |

IN SMOTE-NC, the distance between i1-i2 is calculated as follows:

The standard deviation of the three continuous features (C1, C2, C3) associated with minority class min, are 70.71, 31.82, 21.92 respectively and the median m of these numbers is 31.82.

Distance $_{\text{SMOTE-NC (i2 - i1)}}$ = $\sqrt{(100-200)^2 + (20-65)^2 + (85-54)^2 + (31.82)^2}$

$= 118.31$

In SMOTE-ENC we do the encoding of nominal labels as follows:

Imbalance Ratio (ir) = 2/5 = 0.4

Total a label in our training dataset (e) = 2

Total a label associated with minority class (min) o = 1

If a label had equal affinity towards min and maj class, then total number of a labelled instances in minority samples would be e` = e ∗ ir = 2 x 0.4 = 0.8

Encoding of a for all target min = $\frac{(o-e`)}{e`} * m$ = $\frac{(1-0.8)}{0.8} x$ 31.82 = 7.95

Total b label in our training dataset (e) = 3

Total b label associated with minority class (min) o = 1

If b label had equal affinity towards min and maj class, then total number of b labelled instances in minority samples would be e` = e ∗ ir = 3 x 0.4 = 1.2

Encoding of b for all target min = $\frac{(o-e`)}{e`} * m$ = $\frac{(1-1.2)}{1.2} x$ 31.82 = -5.30

Distance $_{\text{SMOTE-ENC (i2-i1)}}$ = $\sqrt{(100-200)^2 + (20-65)^2 + (85-54)^2 + (7.95+5.30)^2}$

$= 114.72$

We can see that the distance between two data rows in our sample dataset is reduced when calculated by SMOTE-ENC. Hence, it can be said that, in SMOTE-NC, inter and intra level distance of categorical levels are always the same; whereas, in SMOTE-ENC, nominal features are encoded as numeric values and difference between two such numeric value reflects the amount of change of association with minority class and hence seems to produce better result than SMOTE-NC.

By applying this algorithm on the nominal features, we encode each label as a numeric value; where higher value of a label indicates stronger association with minority class. This encoding is inspired from Pearson's chi-squared test[25] which can confirm if two categorical variables are related. However, in this method, our objective is not to determine if the nominal features of the samples are related to the target, rather to quantify the distance between two data points when label of a categorical variable is changed. Hence, we do not use chi-squared test, i.e., χ2, rather use chi (χ) for each label of nominal feature to measure amount of association of that label with the minority class target. Magnitude of χ (i.e., chi) is a fraction and when included in the distance calculation between two samples can become insignificant as the difference between two values of a continuous feature is often significantly higher. For this reason, if the dataset has continuous features, χ is multiplied by the median of standard deviation of continuous variable to make the value comparable to other continuous attributes. However, if the dataset has only nominal features, chi (χ) is directly used in distance calculation.

We compared our proposed method, SMOTE-ENC, with SMOTE-NC on five different datasets. These datasets have diverse minority to majority ratio which offers a broad range for testing this algorithm's efficiency. Among these datasets forest cover data has 7 classes. We evaluated our model on a pair of classes of the original dataset and discarded the rest of the instances.

The metrics used in our study to evaluate model's performance was precision, recall, F-beta score, area under precision recall curve and area under ROC curve. Accuracy was not used as an evaluation metric in this paper, as accuracy is not able to capture model performance for highly imbalanced dataset. F-beta Score is the weighted average of precision (ratio of correctly predicted positive observations to the total predicted positive observations) and recall (ratio correctly predicted positive observations to the total positive observations).

$$F_\beta = (1 + \beta^2) * \frac{precision * recall}{(\beta^2 * precision) + recall}, \quad (2)$$

Choice of β signifies the weightage on precision and recall. While 0 < β < 1 assigns more weightage on precision, 1 < β < + ∞ assigns more weightage on recall. β = 1 assigns equal weightage on precision and recall and known as F1-score. ROC is a probability curve and area under ROC curve represents the degree of separability. It identifies how well the model can differentiate between classes as it balances between precision and recall. As the AUC gets higher, the model gives better prediction. This curve is plotted with True Positive Rate/recall against the False Positive Rate. However, ROC plot can be misleading when applied to heavily imbalanced datasets as adding a lot of majority class observations can improve the AUC significantly without any improvement in precision and recall of minority class instances. For imbalanced class problem, precision-recall curve offers better insight[26]. PRC shows the relationship between precision and recall, and its baseline moves with class distribution, e.g., for a dataset with imbalance ratio 1:10, the baseline moves to 0.09, i.e., 1/ (1+10).

Stratified cross-validation was performed during over-sampling to resolve over-optimization problem[27]. Random forest was used as the classification algorithm on the resampled data. We created a combined estimator by using pipeline to be applied on the data which consists of the over-sampling method and the random forest classifier.

In this study, we used two-tailed t-test to show that the improved results obtained using our proposed over-sampling method and the model using existing over-sampling method are statistically significant. The statistical test performed on precision, recall and f1-score results. The p-value of 0.05 has been used as a threshold for the statistical significance.

## 3. Results

We compared SMOTE-ENC performance with SMOTE-NC precision, recall, F1-score, area under PRC and ROC on 5 different datasets. Performance of each one discussed below.

**Table 2.** Comparison between SMOTE-ENC (new method) and SMOTE-NC on minority class data points at threshold 0.5.

| Dataset | IR[1] | N[2] | C[3] | Sampling method | Precision | Recall | F1-score |
|---|---|---|---|---|---|---|---|
| Banking tele-marketing dataset[28] | 9:1 | 12 | 5 | SMOTE-ENC | 31.95 | 46.84 | 37.99 |
|  |  |  |  | SMOTE-NC | 27.33 | 50.40 | 35.45 |
| Credit card dataset[29] | 12:1 | 5 | 14 | SMOTE-ENC | 65.61 | 81.75 | 72.79 |
|  |  |  |  | SMOTE-NC | 59.78 | 84.92 | 70.16 |
| Car dataset[30] | 3:1 | 6 | 0 | SMOTE-ENC | 87.91 | 79.21 | 83.33 |
|  |  |  |  | SMOTE-NC | NA[4] | NA[4] | NA[4] |
| Forest cover dataset (class 2 and 6)[31] | 17:1 | 2 | 12 | SMOTE-ENC | 74.76 | 99.84 | 85.50 |
|  |  |  |  | SMOTE-NC | 74.76 | 99.82 | 85.49 |
| Rain in Australia dataset[32] | 4:1 | 3 | 13 | SMOTE-ENC | 58.45 | 72.24 | 64.62 |
|  |  |  |  | SMOTE-NC | 58.30 | 72.51 | 64.63 |

[1]IR (imbalance ratio) = Total number of samples in training set/Number of minority class samples.
[2]N = Number of nominal features in the dataset. [3]C = Number of continuous features in the dataset.
[4]NA = SMOTE-NC cannot be applied on this dataset.

### 3.1. Evaluation on banking tele-marketing dataset

This is a publicly available dataset to predict the success of tele-marketing

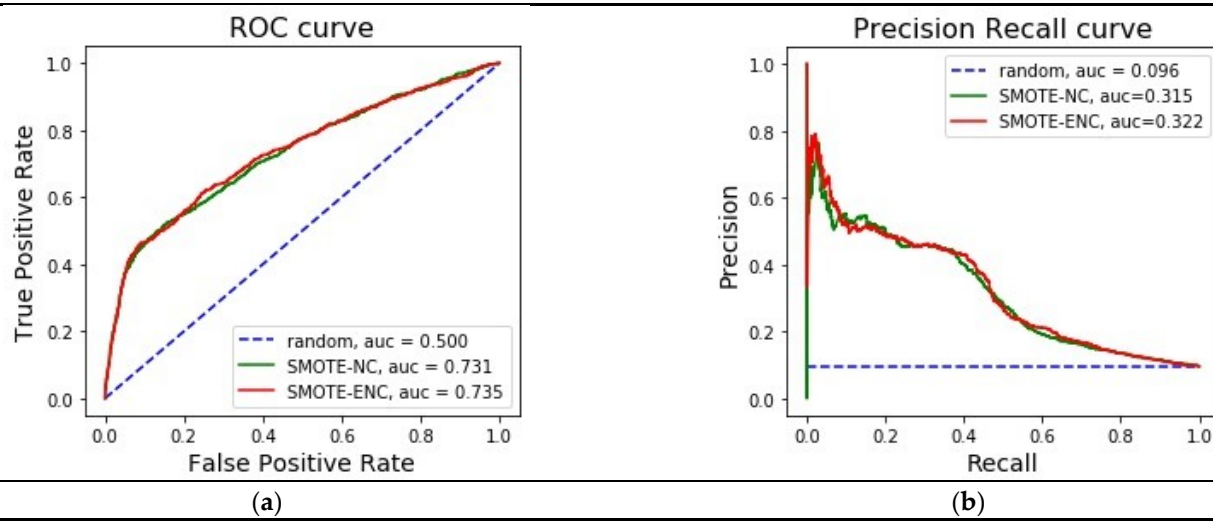

**Figure 1.** (**a**) Comparison between SMOTE-ENC and SMOTE-NC on banking dataset by area under ROC curve; (**b**) Comparison between SMOTE-ENC and SMOTE-NC on banking dataset by area under PR curve.

From Table 2, it is observed that, for bank dataset, SMOTE-ENC achieved significant improvement in precision and F1-score at threshold 0.5 than SMOTE-NC on minority class instances. We further compared these two methods by ROC and precision-recall curve. Figure 1 demonstrate that SMOTE-ENC is a better sampling method than SMOTE-NC for banking dataset when identifying minority class instances is the focus of interest. We further looked at the statistical significance of our improved results and identified that the p-value from two tailed t-test was less than 0.002 for precision, recall and f1-score. This shows that our improved performance is statistically significant. Therefore, it can be said, SMOTE-ENC has been able to interpret nominal features more accurately while generating synthetic data points. We further looked at whether SMOTE-ENC has any effect on the change of importance of features (Figure 2).

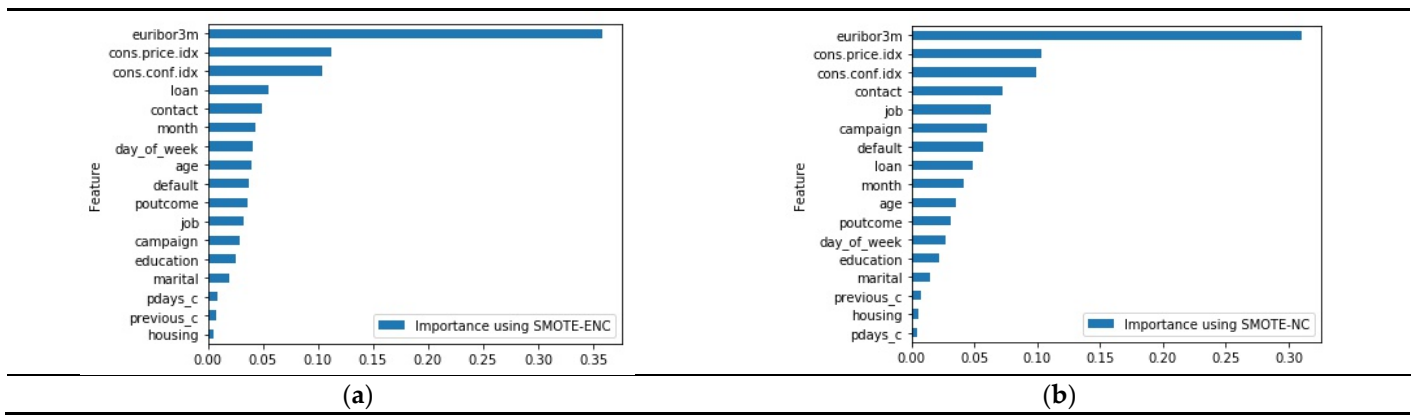

**Figure 2.** (**a**) Feature importance obtained from random forest classifier using SMOTE-ENC; (**b**) Feature importance obtained from random forest classifier using SMOTE-NC.

In the Banking Dataset, euribor3m, cons.price.idx, cons.conf.idx, age and campaign are continuous features. Figure 2 shows that top 3 important features are continuous variables, which remained unchanged, contributing >0.55 importance towards classification. A slight shift in the importance of low-ranked features was observed.

*3.2. Evaluation on credit card dataset*

This is a publicly available dataset to predict churning customers.

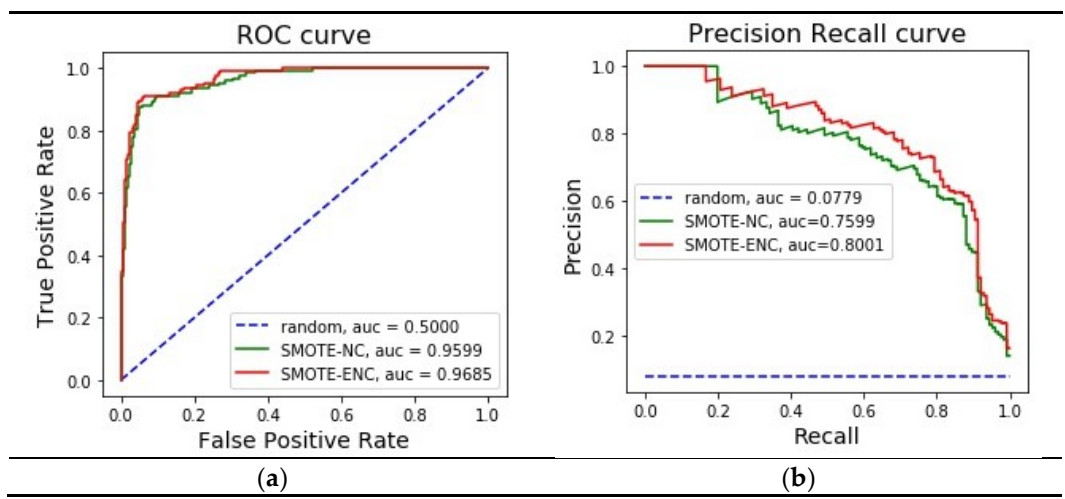

**Figure 3.** (**a**) Comparison between SMOTE-ENC and SMOTE-NC on credit card dataset by area under ROC curve; (**b**) Comparison between SMOTE-ENC and SMOTE-NC on credit dataset by area under PR curve.

Table 2 demonstrates that, SMOTE-ENC yielded significantly better result than SMOTE-NC in terms of precision and F1-score at threshold 0.5 on minority class instances for this dataset and figure 3 shows that SMOTE-ENC attains higher ROC and PRC score than SMOTE-NC. Difference between these two model's performances was statistically significant as p-value from two tailed t-test was less than 0.004 for precision, recall and f1-score. Hence, it can be said that, the new SMOTE-ENC method is the better sampling method than SMOTE-NC for this credit card dataset when identifying minority class instances is the point of interest. We further looked at whether SMOTE-ENC has any effect on the change of importance of features (Figure 4).

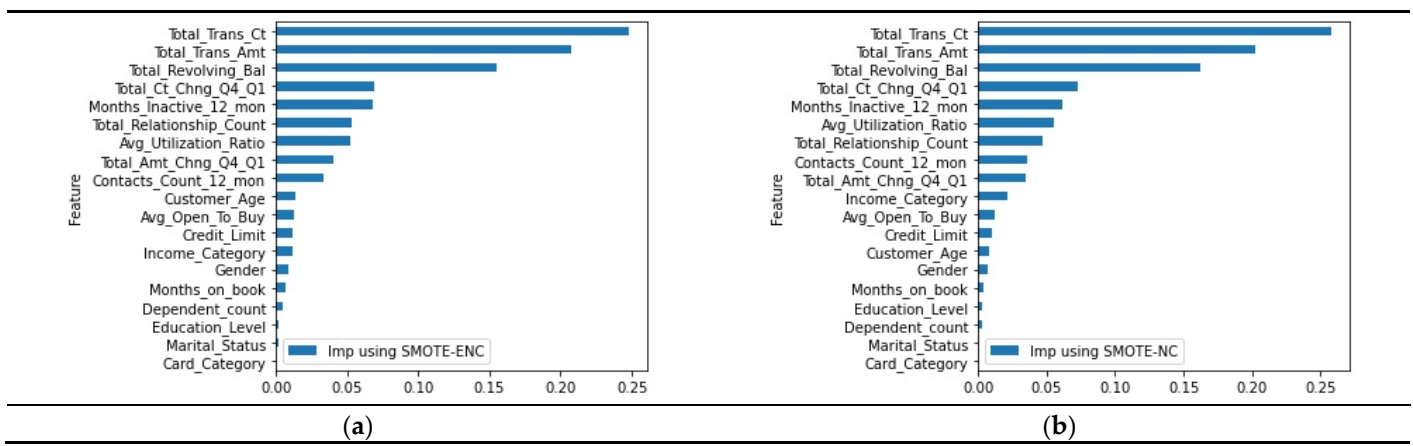

**Figure 4.** (**a**) Feature importance obtained from random forest classifier using SMOTE-ENC; (**b**) Feature importance obtained from random forest classifier using SMOTE-NC.

From Figure 4, it can be observed that, none of the five nominal features of this dataset, came up in the top 10 feature importance ranking using random-forest classifier. Application of our method, SMOTE-ENC, did not change this ranking showing that our SMOTE-ENC improves performance by interpreting nominal features in a different way, however didn't diminish continuous features' contribution in model prediction.

*3.3. Evaluation on car dataset*

This is a publicly available dataset to evaluate cars.

In SMOTE-NC, distance between two levels of a categorical feature is the median of standard deviations of all continuous features for the minority class. So, and in order to SMOTE-NC work, datasets need to have at least one continuous feature.

And, the car dataset, used in this study, does not have any continuous feature to predict the target outcome. All of its' attributes are multi-level categorical features. For this reason, SMOTE-NC cannot handle this dataset.

However, the new SMOTE-ENC method can be applied on this dataset, because, in our SMOTE-ENC algorithm inter-level distance of a nominal feature is not dependent on continuous feature's presence and hence has been generalized to handle both mixed-dataset (i.e., dataset containing nominal and continuous features) and nominal only datasets.

Table 2 shows that, SMOTE-ENC yielded F1-score of 83.33 at threshold 0.5 on minority class instances for this dataset and from figure 5 it can be observed that SMOTE-ENC produced high ROC and PRC score as well.

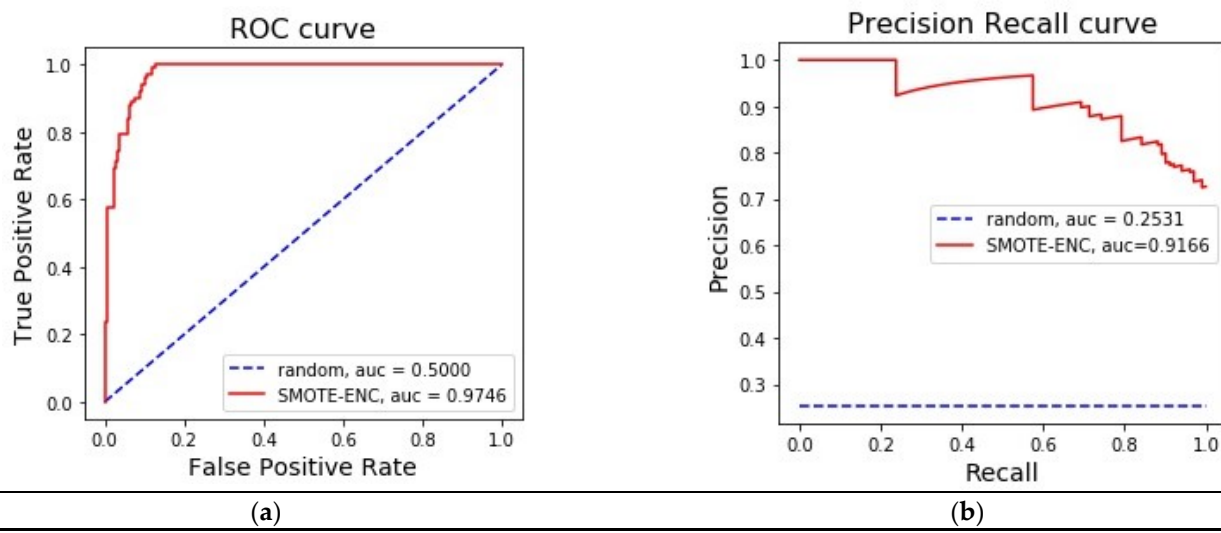

**Figure 5.** (**a**) Application of SMOTE-ENC on car dataset by area under ROC curve; (**b**) Application of SMOTE-ENC on credit dataset by area under PR curve.

*3.4. Evaluation on forest cover dataset with class 2 and 6*

This is a publicly available dataset to predict forest cover type

For Forest cover dataset with class 2 and 6, it was observed from Table 2 that, while precision was consistent for both of the sampling technique, using SMOTE-ENC increased recall of the minority class observations, which resulted in marginal improvement of F1-score for SMOTE-ENC at threshold 0.5.

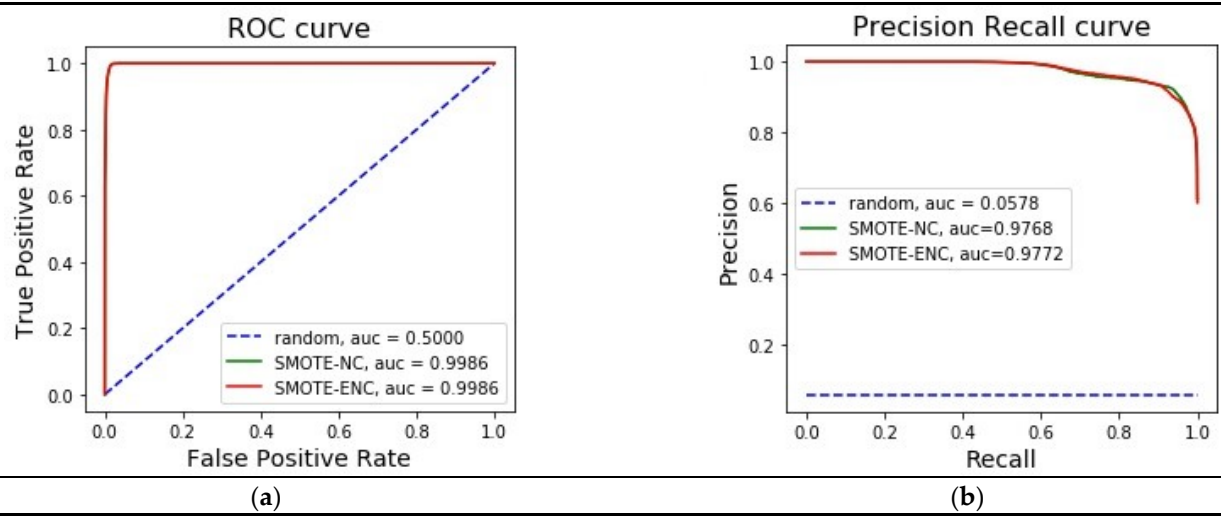

**Figure 6.** (**a**) Comparison between SMOTE-ENC and SMOTE-NC on forest cover dataset with class 2 and 6 by area under ROC curve; (**b**) Comparison between SMOTE-ENC and SMOTE-NC on forest cover dataset with class 2 and 6 by area under PR curve.

The precision-recall curve (Figure 6) showed that SMOTE-ENC was marginally better than SMOTE-NC for this dataset. However, the area under ROC and PR curve being high implies that the classes are easily separable, which makes the importance of sampling method less prominent. For this reason, we one-hot encoded the categorical features and applied SMOTE to resample the data. In this dataset, we had 2 categorical features, soil type and wilderness_area. While the former contained 40 labels, the latter had 4 labels.

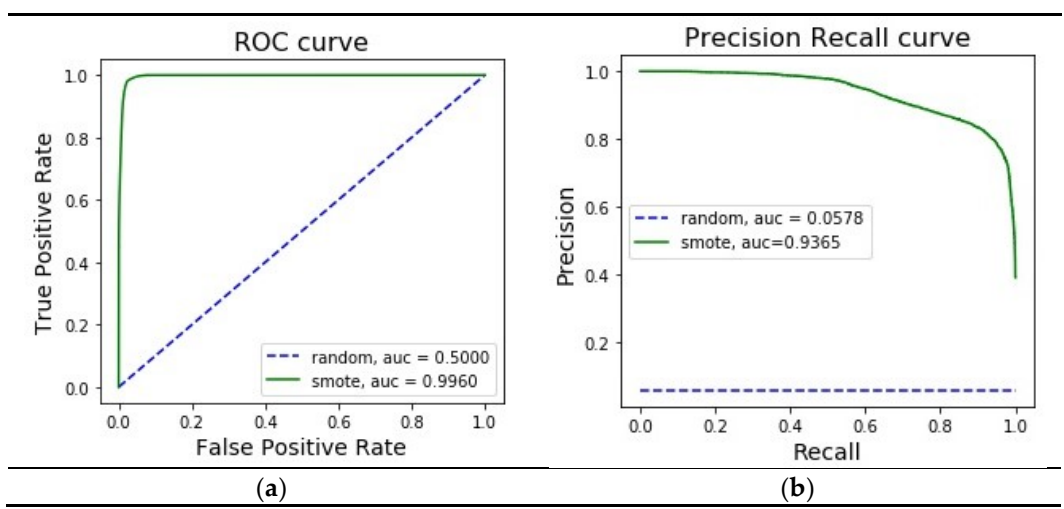

**Figure 7.** (**a**) ROC curve for forest cover dataset with class 2 and 6 using SMOTE; (**b**) PR curve for forest cover dataset with class 2 and 6 using SMOTE.

From Figure 7, it was observed that, one-hot encoding of multi-label features decreased area under ROC and PR curve. Thus, the importance of interpreting nominal features in a different way than interpreting continuous features in over-sampling method can be ascertained. We further showed that importance of categorical features (soil type and wilderness_area) was picked up by SMOTE-ENC and SMOTE-NC but not by SMOTE (Figure 8).

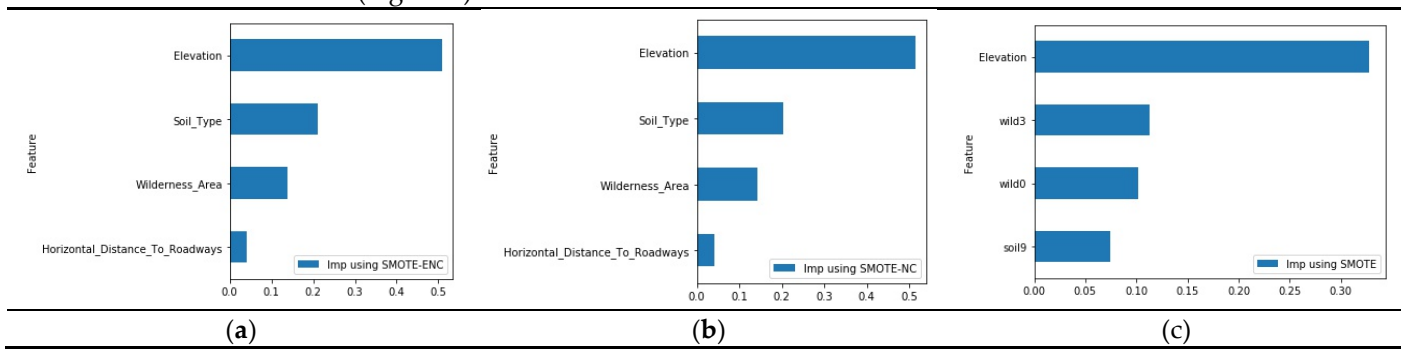

**Figure 8.** (**a**) Top 4 important features of random-forest using SMOTE-ENC; (**b**) Top 4 important features of random-forest using SMOTE-NC; (**c**) Top 4 important features of random-forest using SMOTE.

*3.5. Evaluation on rain dataset*

This is a publicly available dataset to predict occurrence of rainfall on the next day based on today's weather attributes.

SMOTE-ENC provides better precision whilst reducing recall (Table 2). The magnitude of reduction in recall performance being higher than the magnitude of improvement in precision, the harmonic-mean of precision and recall – the F1 score reduces.

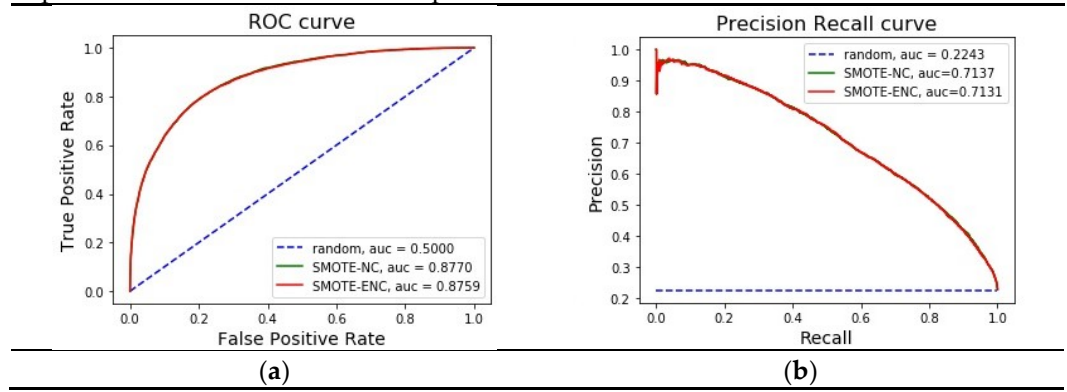

(a) (b)

**Figure 9.** (**a**) Comparison between SMOTE-ENC and SMOTE-NC on rain dataset by area under ROC curve; (**b**) Comparison between SMOTE-ENC and SMOTE-NC on rain dataset by area under PR curve.

ROC and Precision-Recall curves (Figure 9) showed that SMOTE-NC yields marginally better performance than SMOTE-ENC.

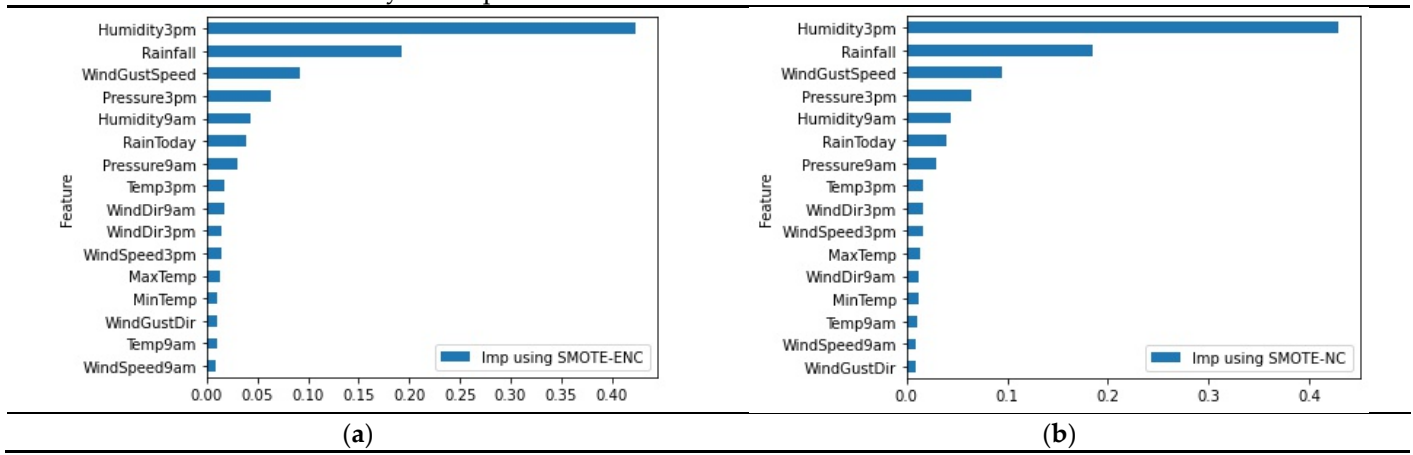

(a) (b)

**Figure 10.** (**a**) Feature importance of random-forest using SMOTE-ENC; (**b**) Feature importance of random-forest using SMOTE-NC.

In this dataset, there are only 3 nominal features – WindGustDir, WindDir9am and WindDir3pm and from the Figure 10, it can be observed that, none of the three features came up in the top 5 feature importance ranking using random-forest classifier. Application of our method, SMOTE-ENC, did not change this ranking showing that our SMOTE-ENC improves performance when there is a reasonable amount of association between nominal feature and the class outcome.

## 4. Discussion

SMOTE and its different variant methods, including Borderline-SMOTE, ADASYN, SMOTE-Tomek etc. only deals with continuous features. However, in real-world, datasets often consist both continuous and categorical features. One way to use variants of these SMOTE techniques on such datasets is by one-hot encoding categorical features first and then applying over-sampling method on them. However, it generates two major problem:

firstly, in case the dataset contains several multi-label nominal features, one-hot encoding increases data dimension significantly, which in turn brings curse of dimensionality[33]. Secondly, being unaware of nominal features, these SMOTE variants produce multiple new labels of these attributes in the synthetically generated resampled data, which does not have any physical significance. SMOTE-NC resolves these problems, but each label is considered to have equal propensity towards target class, muddling the feature's contribution towards distance calculation. Whereas, our developed variation, SMOTE-ENC, differentiates label of association between labels of a particular nominal feature, capturing nominal feature's contribution in distance calculation more precisely. E.g., let's suppose, there is a dataset which contains two multi-level categorical features (c1 and c2) along with few continuous features. In SMOTE-NC algorithm, inter-level distance of c1 feature will be same as the inter-level distance of c2 feature. Because, in SMOTE-NC, distance between any two levels of the nominal attributes is the median of standard deviations of the continuous features of the data. In other words, inter-level distance of nominal feature is not dependent on that feature, rather on the continuous attributes of the dataset. Whereas, in our version of SMOTE algorithm, inter-level distance of c1 feature will be different than that of c2 feature and the distance will not be dependent on the continuous features, rather on that particular nominal feature's distribution. Hence, if inter-level distance of c1 is more than that of c2, c1 will have higher contribution while calculating distance between two data points.

From ROC and precision-recall curves in the Results section, we can see that, of 5 datasets, SMOTE-ENC yielded better result in 4 data. In these cases, when resampled data was applied on random-forest classifier, nominal features turned out to be significant drivers. From this observation, it can be inferred that, SMOTE-ENC is likely to yield better result when there is some association between the categorical features and the target class. When, the association is poor, SMOTE-ENC is not able to yield better result than SMOTE-NC.

From the Table 2, it can be observed that, even when ROC and precision-recall curve shows SMOTE-ENC and SMOTE-NC to yield similar performance, there is a difference between these methods' performance in terms of precision and recall. While for banking, credit card and rain datasets, SMOTE-ENC generates better precision, for forest cover dataset, it produces improved recall. Thus, SMOTE-ENC and SMOTE-NC can be differentiated when either of precision and recall gets higher priority and $F_\beta$ score is the better evaluator than F1-score.

From Figure 1 and figure 3, it is also evident that, SMOTE-ENC out-performed SMOTE-NC method substantially on banking and on credit card dataset. For rest of the datasets, the new algorithm could not attain significant improvement. It was observed that, in these two datasets (Table 2), proportion of number of nominal features to total number of features is more than 25%, whereas, for rest of the datasets this proportion is less than 20%. From this observation, it can be inferred that, in these two datasets, a change in interpretation of the nominal features can bring a considerable amount of change in interpretation of whole dataset. This might be the reason why SMOTE-ENC was able to achieve substantial improvement on these datasets.

When a dataset has one or more multi-label nominal features, SMOTE-NC fails to identify if one or few labels are more associated towards minority class instances than others and thus muddles the significance of the feature in distance calculation. In our method, SMOTE-ENC, when the data has multiple nominal features, for each nominal feature the distance between its labels are calculated in isolation, thus the inter-label distance for each nominal feature is different. Whereas, in SMOTE-NC, the inter-label distance between every nominal feature is considered same.

From table 2, it can also be observed that SMOTE-NC could not be applied on the car dataset. This car dataset, used in this study, does not have any continuous feature but only multi-level categorical features. And SMOTE-NC algorithm can only function when there is at least one continuous feature. In SMOTE-NC algorithm distance between two levels of a nominal feature is the median of standard deviations of all continuous features for the minority class. Therefore, if the dataset does not have any continuous feature, median of standard deviations of the continuous features cannot be calculated and the algorithm fails to generate synthetic data points to balance the dataset. Whereas, our new SMOTE-ENC algorithm, has been generalized to handle both mixed (i.e., dataset with continuous and categorical features) and categorical-only datasets.

## 5. Conclusion

From the experimental results, it can be inferred that the proposed method over-performs the existing method when the dataset has a substantial number of categorical features. SMOTE-ENC is an extension of SMOTE and can be enhanced further and implemented with other variants of SMOTE, e.g., SVMSMOTE, ADASYN, SMOTEENN, SMOTE Tomek etc. This algorithm is equipped to handle datasets with only nominal features as well, which is not possible in the existing SMOTE-NC method. SMOTE-ENC implements one way to interpret the relationship between nominal feature and target class by encoding each label as $\chi$. There are various other ways to calculate amount of association between two nominal features and those can be implemented and evaluated on skewed dataset in future.

**Author Contributions:** Conceptualization, Mimi Mukherjee; methodology, Mimi Mukherjee and Matloob Khushi; software, Mimi Mukherjee; validation, Mimi Mukherjee and Matloob Khushi; formal analysis, Mimi Mukherjee; investigation, Mimi Mukherjee and Matloob Khushi; data curation, Mimi Mukherjee; writing—original draft preparation, Mimi Mukherjee; writing—review and editing, Matloob Khushi; visualization, Mimi Mukherjee; supervision, Matloob Khushi. All authors have read and agreed to the published version of the manuscript.

**Funding:** This research received no external funding.

**Data Availability Statement:**

- The data presented in this study are openly available in [Moro, S., P. Cortez, and P.J.D.S.S. Rita, A data-driven approach to predict the success of bank telemarketing. Decision Support Systems, 2014. 62: p. 22-31 DOI: https://doi.org/10.1016/j.dss.2014.03.001].
- Publicly available datasets were also analyzed in this study. This data can be found here: [Asuncion, A. and D. Newman, UCI machine learning repository. 2007], [Young, J. and Adam young, Rain Dataset: Commonwealth of Australia 2010, Bureau of Meteorology. 2018], [Sakshi, G., Credit Card customers - Predict Churning customers. 2020] and at [Dua, D.a.G., C, [31] Machine Learning Repository. 2017].

**Conflicts of Interest:** The authors declare no conflict of interest.